\documentclass{sigkddExp}
\usepackage{graphicx}
\usepackage{amsmath}
\usepackage[caption=false]{subfig}
\graphicspath{ {images/} }
\usepackage[labelfont=bf,textfont=bf]{caption}
\usepackage{makecell}
\usepackage{flushend}
\usepackage{cite}
\usepackage{booktabs}
\usepackage{multirow}
\usepackage{siunitx}
\usepackage{colortbl,dcolumn}
\usepackage{hyperref}

\definecolor{green2}{rgb}{.23,.59,.17}
\definecolor{green4}{rgb}{.17,.73,.09}
\definecolor{green1}{rgb}{0.169,0.51,0.316}
\definecolor{green3}{rgb}{0.075,0.6236,0.301}

\numberofauthors{4}
\author{Sajan Kedia\\Myntra Designs, India\\sajan.kedia@myntra.com
\and Manchit Madan\\Myntra Designs, India\\manchit.madan@myntra.com 
\and Sumit Borar\\Google, India\\sumitborar@gmail.com\thanks{Work done while at Myntra}}

\title{Early Bird Catches the Worm: Predicting Returns Even Before Purchase in Fashion E-commerce}

\begin{document}

\maketitle

\begin{abstract}

With the rapid growth in fashion e-commerce and customer-friendly product return policies, the cost to handle returned products has become a significant challenge. E-tailers incur huge losses in terms of reverse logistics costs, liquidation cost due to damaged returns or fraudulent behavior. Accurate prediction of product returns prior to order placement can be critical for companies. It can facilitate e-tailers to take preemptive measures even before the order is placed, hence reducing overall returns. Furthermore, finding return probability for millions of customers at the cart page in real-time can be difficult.

To address this problem we propose a novel approach based on Deep Neural Network. Users' taste \& products' latent hidden features were captured using product embeddings based on Bayesian Personalized Ranking (BPR). Another set of embeddings was used which captured users' body shape and size by using skip-gram based model. The deep neural network incorporates these embeddings along with the engineered features to predict return probability. Using this return probability, several live experiments were conducted on one of the major fashion e-commerce platform in order to reduce overall returns (discussed in detail in Section 5).  

\end{abstract}


\section{Introduction}
E-commerce sector is the fastest growing sector, with an expectation to reach \$4 trillion by 2020\footnote{\url{https://www.huffingtonpost.com/michael-lazar/retailers-people-want-eas_b_12759542.html}}. In today's competitive market, to increase customer experience, most companies are coming up with hassle-free return policies\cite{bonifield2010product}. Customers are usually allowed a return within a month or a similar time period. This policy\cite{article} has improved customer engagement, revenue, purchase rate, customer experience, and repeat buying behavior. However, with the promise of easy returns also comes the inevitable high return rate. Studies have shown that one-third online orders are returned\footnote{\url{https://www.wsj.com/articles/rampant-returns-plague-eretailers-1387752786}}. Similar behavior has been observed on a major fashion e-commerce platform, with a major reason for returns being the size and fit.

With high return rates comes the problem of huge reverse logistic cost. Since the return policy is lenient, customer return the products after using or damaging\cite{doi:10.1108/03090561011032694} them to a point that does not meet the quality norms of re-selling the product\cite{Chen2009TheIO}. Products won't be live during the return period resulting in lost-sales. The “last mile” delivery chain is the most energy-intensive task which impacts the sustainability of e-commerce, both economically and ecologically\cite{doi:10.1108/09600031011018055}. Hence, returns impact many segments of fashion business such as customer experience, supply chain management, call center demand, inventory, and customer service. It eats a major share of the profit margin of e-tailers. 

Predominantly, the industry is trying to solve this problem post-order, i.e once the order is placed. Most of the existing literature\cite{ma2016predictive}\cite{toktay2001forecasting}\cite{7385772} talks about forecasting returns to solve operational issues. Predicting customer return chances in advance during browsing or at shopping cart page will empower e-tailers to prevent such orders. In this paper, we propose a novel deep neural network based approach to predict customer's likelihood of return even before an order is placed. To get users' taste and products' latent hidden features, Matrix Factorization(MF) based BPR\cite{rendle2009bpr} model is used to detect similar products in a cart. In fashion industry, most of the returns are due to size \& fit issues\footnote{\url{https://www.shopify.com/enterprise/ecommerce-returns}}. To capture this in our model, we have created user's sizing vector using skip-gram based model\cite{DBLP:journals/corr/abs-1301-3781}. In our network, we have used both these vectors along with engineered features. A hybrid dual-model approach is proposed to first, predict the return probability at the cart level, and second, predict at an individual product level. Prediction happens in real-time at the cart page, so that preemptive actions can be taken based on the return probability value. The possible set of actions are:

\begin{itemize}
    \item Personalized Shipping charges
    \item Make product non-returnable by giving an additional coupon
    \item Try \& Buy options
    \item In case of return related refund, money goes directly to wallet which can only be used for shopping again on the same platform
    \item Restricting payment options like Cash on delivery
    \item Advance alert for reverse logistics
    \item Artificially show the product as out of stock \& prevent the user from placing that order\cite{Urbanke2015PredictingPR}
\end{itemize}

Most of these action items require the return prediction at a cart level whereas for rest it should be at an individual product level. The proposed hybrid dual-model serves both the scenarios. As shown in Figure 1, the first two images on the left side demonstrate the personalized delivery charge experiment. One has free delivery charge whereas other has 149 as a delivery charge based on the return probability of that cart. In the middle image, all the products are similar which can be a reason for high return probability as compared to the leftmost image where all items are different. The rightmost image demonstrates the non-returnable experiment at the product level. For products having a higher probability of return, an option of making it non-returnable is given to the customer by providing an additional coupon. This experiment helped in reducing returns \& reverse logistics. Here, the model works at an individual product level instead of cart level.

The rest of the paper is organized as follows. In Section 2, we briefly discuss the related work. We introduce the Methodology in Section 3, that comprises of feature creation and explanation of hybrid dual-model. In Section 4, Results and Analysis are discussed. Then we present Experiment results in Section 5 and conclude the paper in Section 6.

\begin{figure}[h]
\centering
\includegraphics[height=2.2in, width=3.5in]{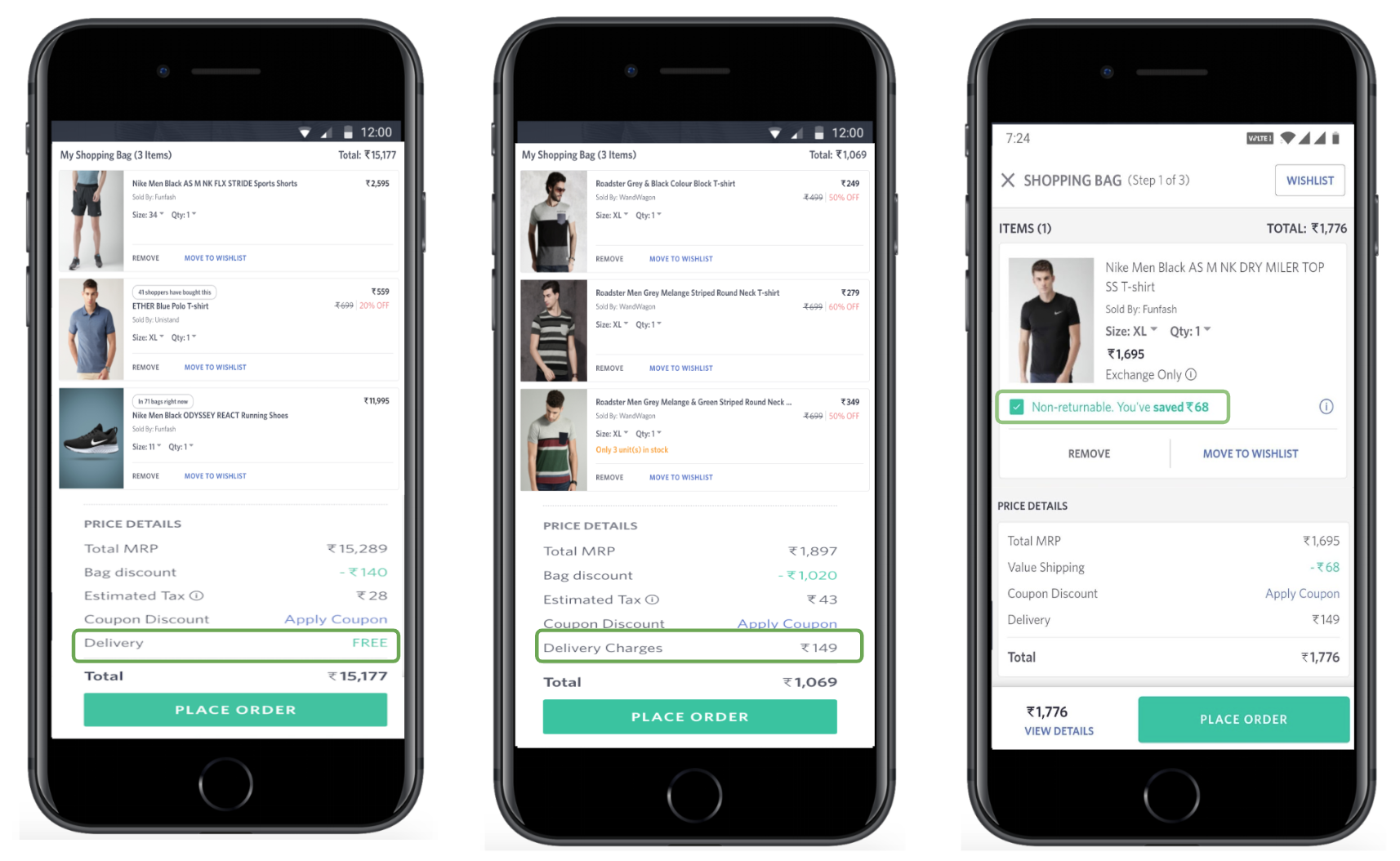}
\caption{First two images from the left demonstrate personalized delivery charges for the same user having different carts. The third image shows a non-returnable option by giving a small additional coupon for the product having high return probability}
\label{figure:action_items}
\end{figure}

\section{Related Work}

In this section, we have briefly reviewed the existing literature on return prediction. In \cite{Urbanke2015PredictingPR} authors used Mahalanobis Feature Extraction to predict product returns in e-commerce. Here a prediction can only be made once the user has placed an order, due to which e-tailer can't take preventive measures to reduce the returns. Also, this is not designed for user-product level prediction.
In \cite{toktay2001forecasting}, time-series methods are used to forecast return volume.
In \cite{DBLP:conf/kdd/LiHZ18},  authors are trying to predict returns using a weighted hybrid graph to represent the rich information in the product purchase and return history.  \cite{ma2016predictive} talks about forecasting the return quantity \& time using the historical sales data which is used to take inventory management decisions. Predicting accurately the number of returns won't be helpful in adopting preemptive measures since the prediction is not done during the order placement. To address these issues we are proposing real-time return prediction at cart page even before the customer places the order. Our prediction being at a user-product level is personalized for all users- at different cart (basket) level.

\section{Methodology} \label{broad-narrow}
Given a user $u$ and current cart $C$ which can contain multiple products, our objective is to come up with return probability score denoted by $f(u,C)$ for a user \& cart combination. This probability score is used to implement the action items mentioned in Section 1 to reduce the overall returns. Analysis showed that number of similar items in a cart impacts this score. Hence, we computed:

\begin{center}
$g(P_i, P_j)$ representing similarity between product $P_i$ and $P_j$
\end{center}

    Since we don't have any explicit ratings available for the products, the only way to find similarity is by using implicit signals from user and product interaction graph. For this, we've used the Matrix Factorization\cite{mnih2008probabilistic} based Bayesian Personalized Ranking (BPR) approach. 

	Another significant cause of returns in fashion industry is due to size \& fit issues\cite{arora2016decoding}. It's very difficult to solve this problem since most brands have a different fitting for the same size number. To address this problem we have created sizing vectors which are personalized at a user, brand level based on their historical data. Skip-gram based model is used to create sizing latent vectors. Finally, a deep neural network model is trained using these embeddings along with the engineered features to compute $f(u,C)$.

	We propose a hybrid dual-model approach. First, a deep neural network model predicts $f(u,C)$ for the whole cart. Second, gradient boosted techniques are used to predict the return probability at a product level.  

\subsection{Bayesian Personalized Ranking (Matrix Factorization) Embedding} \label{word2vec-embeddings}
Being one of the largest players in fashion e-commerce, there are approximately 600K products live on our platform at any given time. Typically, 20\% of the products lead to around 80\% of the revenue, hence it's very difficult to get signals for most of the long-tail products. To solve this problem, product embeddings are created using Matrix Factorization (MF) approach\cite{hoyer2004non}.

To apply MF, user-product interaction matrix is required. Due to the absence of explicit ratings of products, implicit signals are generated using features like the number of views, clicks, cart, order from the user clickstream data. Implicit ratings are calculated by the weighted sum of these signals which are learned by training a classifier. This gives user- product interaction matrix as shown in Figure 2.

To find product embeddings, popular MF based approach- BPR is used. The algorithm works by transforming the user-product interaction matrix into lower dimensional latent vectors which capture the hidden attributes of the products.
Bayesian Personalized Ranking (BPR)\cite{rendle2009bpr} works on pairwise ranking. Loss function of BPR:

\begin{equation}
   -\sum_{(u,i,j)}{}{\ln \sigma(x_{uij}) + \lambda_\Theta || \Theta ||^2}
\end{equation}

where $u,i,j$ are the triplets of product pairs $(i,j)$ and user $u$ available in the interactions dataset. It represents user likes product $i$ over product $j$. And, $x_{uij}$ denotes the difference of affinity scores of the user $u$ for product $i$ and product $j$. $\Theta$ are the model parameters and $\lambda_\Theta$ is model specific regularization parameter.

\begin{center}
    $g(P_i, P_j)$ is computed using BPR, where $g(P_i, P_j)$ represents similarity between product $P_i$ and $P_j$
\end{center}

Product embeddings for all the products are generated using MF-BPR. 

\begin{figure}[h]
\centering
\includegraphics[height=2in, width=3.2in]{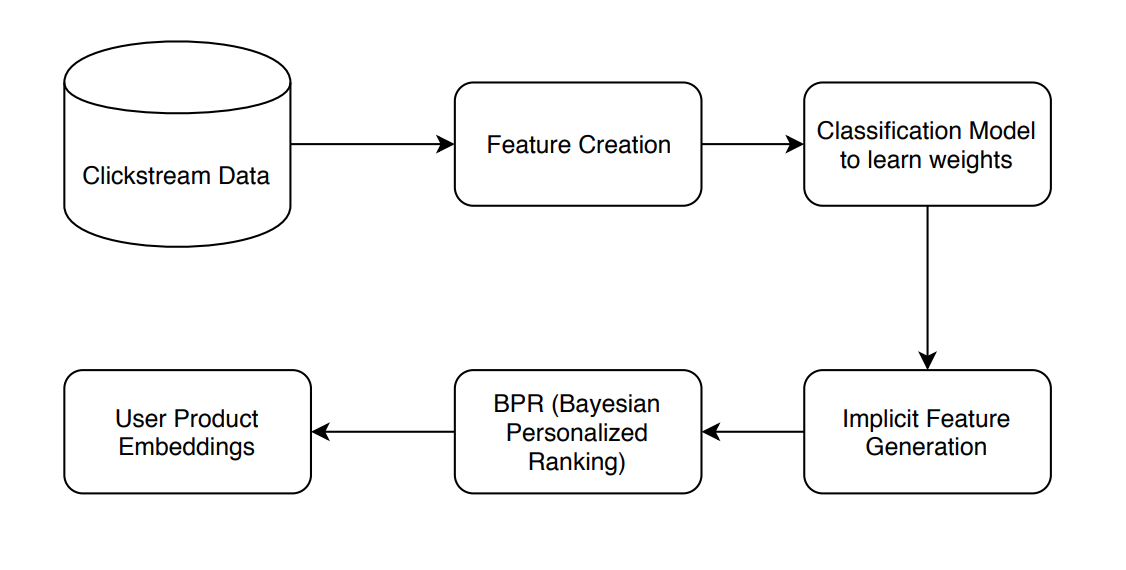}
\caption{BPR-MF Architecture}
\label{figure:architecture1}
\end{figure}

\subsection{Personalized Sizing latent features: Skip-gram approach} \label{word2vec-embeddings}
Size and fit is one of the biggest reason for returns in fashion. It's difficult to address this problem since all brands have different fit even for the same size number. From a customer's point of view, it's a major challenge to decide the perfect size number for different brands. To address this problem, sizing vectors personalized to each user at a brand level are created. From the lifetime clickstream data, first, the user product interaction matrix is computed. Here products are defined in a detailed manner like \textquotedblleft Nike-Men-Shoes-Sports-10\textquotedblright, where 10 is the size. Along with brand, we take category (shoes, jeans, shirts, etc), \& size which helps in understanding all size related attributes for a product.
\newline
    Skip-gram based model\cite{Mikolov2013DistributedRO} is applied on top of the interaction matrix. From this model, the latent features related to size are learned. The network consists of the matrix $M$ where $M_{i}$ represents all the carts of a user $i$ sorted by cart addition time. Given a sequence ${M_{i1}, M_{i2}, M_{i3},..M_{in}}$, here each $M_{ij} \in P$ where $P$ is the entire platform catalog and $M_{ij}$ is the word representation of the product. The objective of the skip-gram model is to maximize the average log probability-

\begin{equation}
  \frac{1}{mn}\sum_{i=1}^{m} \sum_{j=1}^{n} \sum_{-s\leq k\leq s,k \neq 0} \log p(m_{i,j+k}\mid m_{i,j})
\end{equation}

where $s$ is the hyper-parameter denoting the size of training context. Larger $s$ results in higher accuracy as it gets more training examples but at the cost of higher training time. The basic skip-gram formulation defines $p(m_{j}\mid m_{j+k})$ by using softmax function:

\begin{equation}
    p(m_j \mid m_{j+k}) = \frac{exp ({u_{j},v_{j+k}})}{\sum_{k\in M}exp ({u_k,v_k})}
\end{equation}

where $u$ and $v$ are the input and output vector representations of $m_{ij}$. For each product $c_p \in C$,  activation of the hidden layer form a latent feature vector representation $f_p$.

All these vectors $u_i$ are aggregated to do return prediction at cart page.

    Word2Vec\cite{DBLP:journals/corr/abs-1301-3781} is used for the implementation purpose. It gives us sizing vectors which explain the user's body shape \& fit for different brands \& products.
\subsection{Feature Engineering} \label{word2vec-embeddings}
To capture the return affinity of user in model, extensive feature engineering has been done. These features are broadly segregated into three categories:
\begin{enumerate}
    \item Product-level features
    \begin{itemize}
        \item brand, category, product age
        \item product level return score for different periods of time- last month, quarter, half-year, year.
        \item $n$ style group as the number of same products with different colors in a cart.
        \item $n$ similar products as the number of similar products in a cart. 
    \end{itemize}
    
    \item Cart level features
    \begin{itemize}
        \item cart size, order day (weekday/weekend), order time (morning/evening)
        \item brand, category, delivery city, platform type
    \end{itemize}

    \item User level features
    \begin{itemize}
        \item Based on the lifetime data of all users, features like return count, revenue, order count, quantity, payment mode, purchase frequency, were created
        \item Applied clustering techniques at different levels to design new user features like user persona, engagement level, discount affinity score, customer type, purchase intent
    \end{itemize}

\end{enumerate}

One hot encoder was used to transform categorical features like brand, product category, delivery city, day, platform, user clusters. Only important values of the features were transformed by using 65\textsuperscript{th} percentile as the threshold of their occurrence in the data. This helped in avoiding an explosion of features \& treating the long-tail/sparseness.

\subsection{Hybrid Dual-Model} \label{target-classifier}
A dual-model approach is proposed to first, predict the return probability at the cart level, and second, predict at an individual product level. Since millions of orders are placed every week and these orders usually consist of multiple products, it becomes an onerous task to classify returns accurately at an individual product level. To overcome this issue, a classifier is first trained at a higher level (cart) to classify returnable carts. The second classifier is built on the carts classified as returnable by the first classifier, which predicts return probability at an individual product level. The first classifier is a fully connected deep neural network model (discussed in the next section), and the second one is a gradient boosted classifier. The set of optimal hyper-parameters were number of trees: 250, max\_depth: 7, learning rate: 0.005, metric: [auc, binary\_logloss], number of leaves: 150.

\subsection{Deep Neural Network Model} \label{target-classifier}
A fully connected deep neural network is trained using the product embeddings from the MF-BPR model, sizing vectors learned using skip-gram model and engineered features. As shown in Figure 3, the input layer consists of four major groups of features. Since the model is trained at the cart level, an aggregation of product embeddings \& sizing vectors for all the products contained in that cart is done. Resulting vectors serve as input to the network. In this architecture there are two hidden layers which uses ReLU as the activation function. To perform binary classification task in the last layer softmax function is used. Output is the probability value representing the likelihood of return. Probability values pass through the rule-based threshold logic, in order to act upon the action items during live experimentation. These thresholds are decided by a combination of business logic and precision, recall values. As shown in Figure 6, for different values of threshold, precision \& recall values are different. Hence, if the business wishes to target a very small set of high return probable users with strict actions, the threshold will be high resulting in high precision at the cost of a low recall. For example in personalized delivery charge experiment, return probability value above threshold (0.77) were shown delivery charge.

\begin{figure}[h]
\centering
\includegraphics[height=3in, width=3.7in]{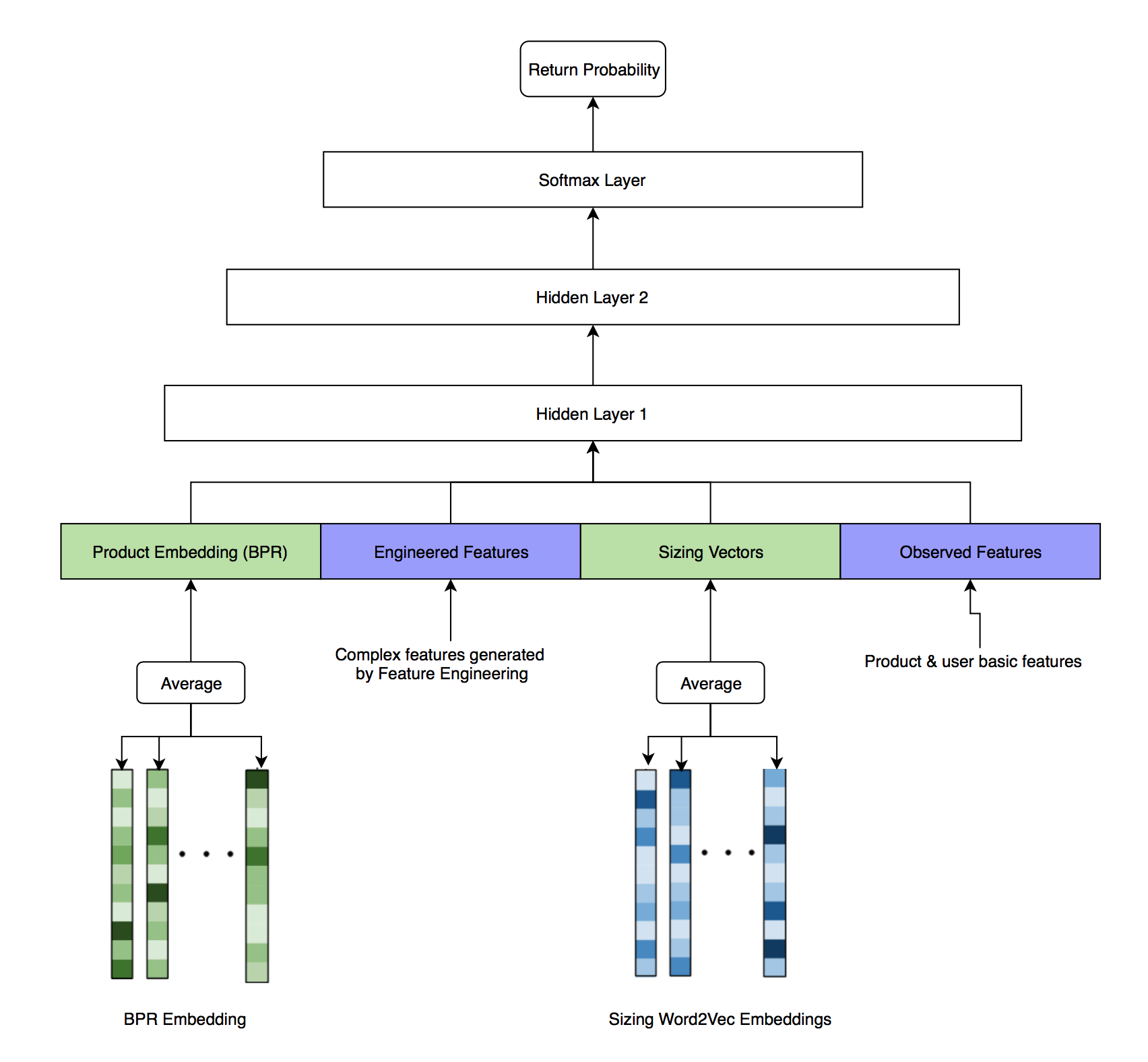}
\caption{A fully connected deep neural network model}
\label{figure:architecture_bpr}
\end{figure}

\subsection{Real-Time Production Architecture}
Real-time preemptive action is the key driver of this algorithm. Hence, whenever a user lands on the cart page, the decision to select the preemptive action should be done in less than 70 milliseconds.
To achieve this, a sophisticated service namely Return Prediction Service (RPS) is designed which takes online, offline features to generate Return Probability. As shown in Figure 4, the real-time architecture is as follows: once a user visits the cart page, the Cart API is triggered which generates the online features (basis the items in the cart) and accumulates the offline features (user activity, product catalogue). Here, online features refer to dynamic features which are created based on products in his cart, example: number of products, number of similar products, revenue, discount, etc. Offline features refer to the set of features which are pre-computed for all users and products, example: user-level features (such as revenue, returned revenue, number of orders, number of products returned, etc) and product-level features (mrp, brand, category, age, etc). All these features are then passed onto the RPS which uses the Trained Model (Deep Neural Network) to give Return Probability. Subject to this Return Probability, appropriate action is taken by the Decision Engine which is then passed back to the Cart API. This real-time architecture empowers us to predict returns at the cart page before an order is placed and take a preemptive action to minimize loss.

\begin{figure}[h]
\centering
\includegraphics[height=2in, width=3.2in]{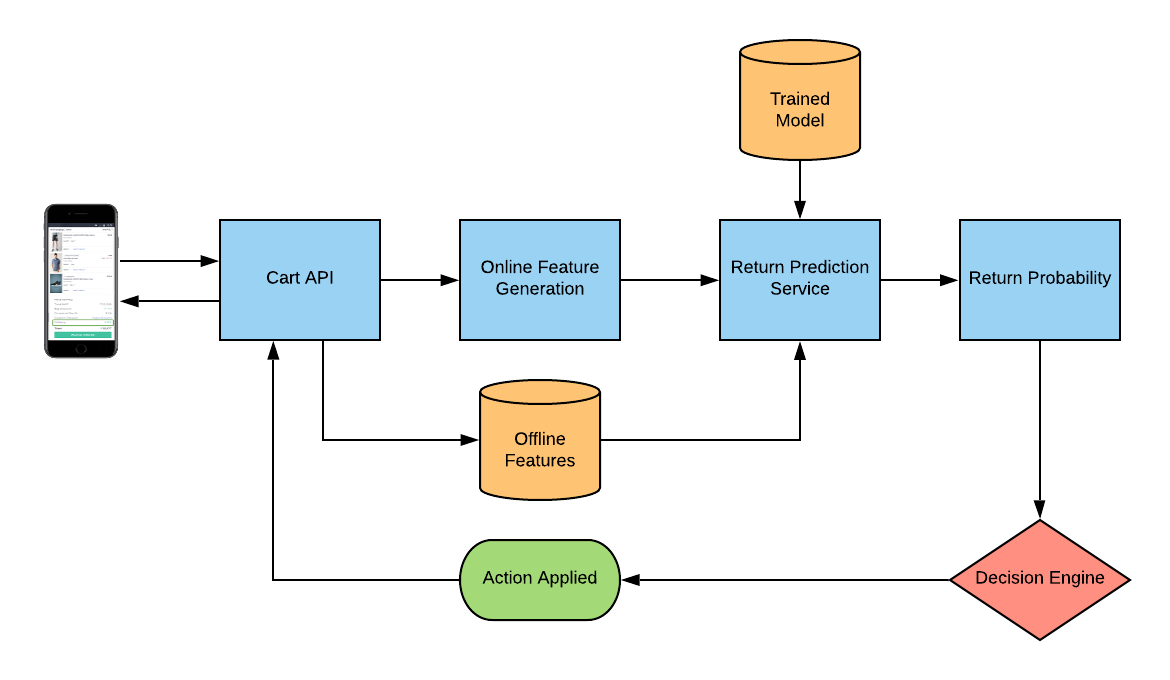}
\caption{Real-time Architecture}
\label{figure:real_time_architecture}
\end{figure}

\section{Results and analysis} \label{word2vec-embeddings}
This section demonstrates the results and analysis of the experiments. First, various data sources used for the modeling purpose are discussed. Thereafter some interesting insights derived from the analysis of the data are presented. Further, the evaluation metrics are discussed. At last, the comparison of different models on key evaluation metrics is done.

\subsection{Data Sources} \label{word2vec-embeddings}

Broadly three categories of data sources are used for the modeling purpose:
\begin{itemize}
\item Clickstream data: It contains all the user activity like clicks, carts, orders 
\item Product Catalogue: Millions of products are cataloged with attributes and details such as color, brand, gender, category, product age, etc
\item Cart data: This contains all the historical carts/basket transactions data
\end{itemize}

\subsection{Analysis} \label{target-classifier}
The analysis shows that out of all the returns 4\% occur when there are some similar products present in the cart. Similar product not necessarily means the same product in different size, color; it can be a similar pattern based on the product's visual attributes. As shown in Figure 1, in the middle image all three t-shirts have similar visual attributes like pockets, design, pattern etc.

\begin{figure}[h]
\centering
\includegraphics[height=2.4in, width=3.3in]{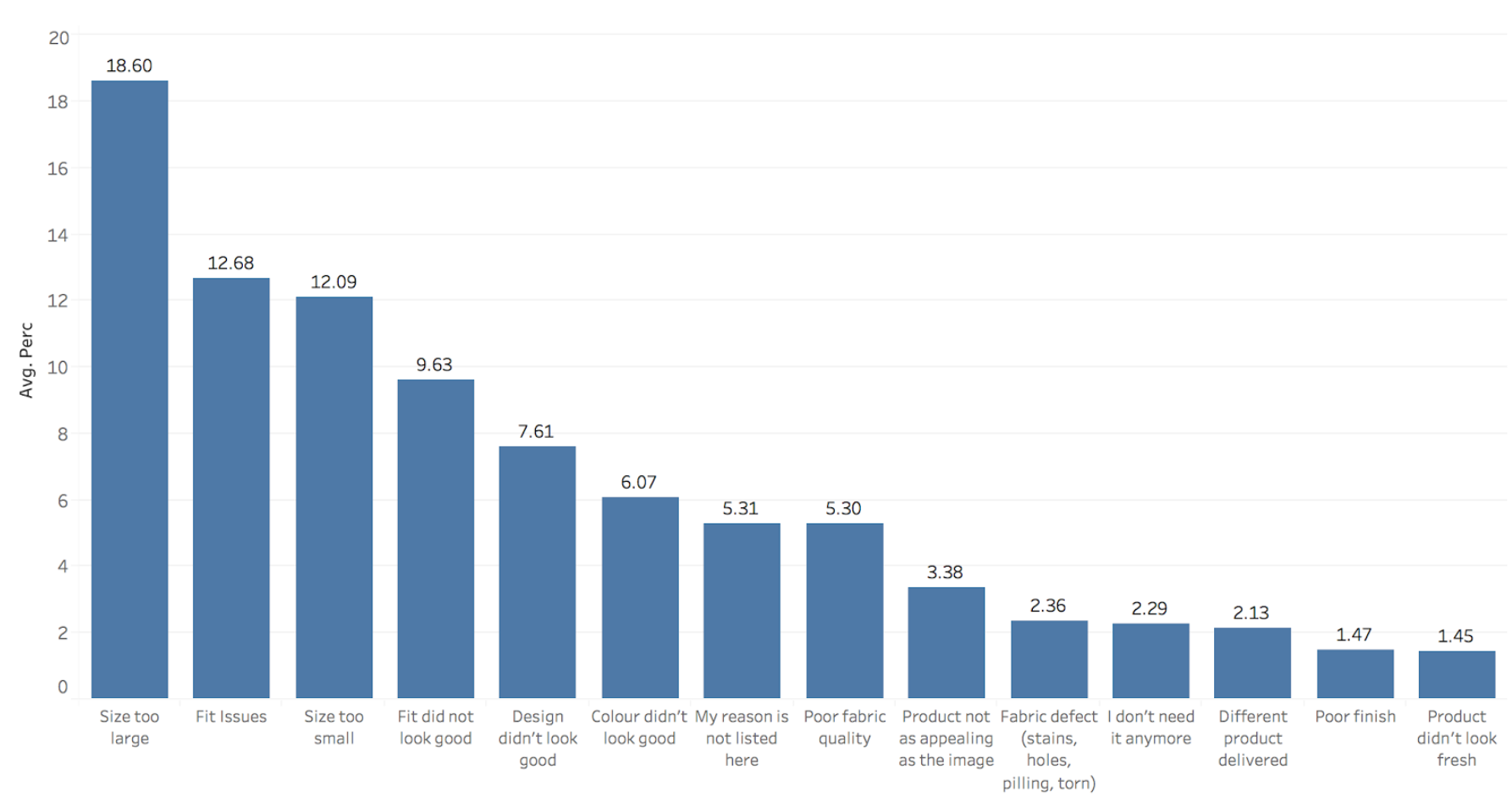}
\caption{Reasons for Return}
\label{figure:architecture1}
\end{figure}

    An order having the same product with different colors accounted for 2\% of the total returned orders. Next analysis captures reasons for returns in online fashion. When a customer returns a product, the reason for return is asked. Figure 5 shows all the major reasons for a return. 53\% of total returns were due to size and fit related issues. This analysis was a major motivation behind creating personalized sizing vectors.

    Return rates are highly dependent on the cart size, with cart size more than five products return rates goes to 72\%, whereas cart with one product has return chances of 9\%. Return behavior change with the day of the week (weekday, week- end), time (morning, evening). Aged inventory had an almost double probability of being returned as compared to newer ones. Historical data showed that the return rate was higher than the platform average for some set of products. These products were 0.16\% of the live catalog and contributed approx 1.9\% returns. The search ranking of these products was de-boosted. Further, some of these were de-listed from the platform to reduce overall returns.

\subsection{Evaluation Metrics} \label{target-classifier}
As this is a classic binary classification problem, standard evaluation metrics are used, namely Precision, Recall, Area under the curve (AUC), Receiver operating characteristic (ROC) curve. Apart from this, during the model deployment in real time, we have used the Precision-Recall curve extensively. As shown in Figure 6, Precision-Recall values change for different values of threshold. This threshold is decided based on the business logic.

\begin{figure}[h]
\centering
\includegraphics[height=2in, width=2.5in]{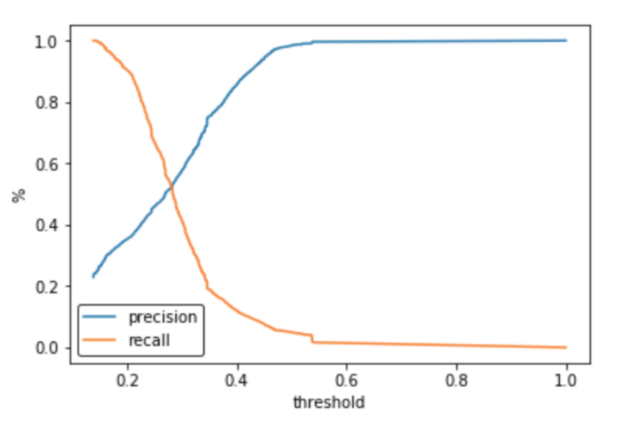}
\caption{Precision Recall value at different threshold}
\label{figure:architecture1}
\end{figure}

If the business wishes to target very high return probable products by providing an additional coupon \& making it non-returnable, the threshold value will be high resulting in high precision and low recall. This threshold value may differ for different experiments. Based on the value of the return probability, we have created segments such as High, Medium, Low representing users' return affinity. Experiments were performed segment-wise, resulting in entirely different A/B tests. For Model performance comparison AUC, precision and recall were used as shown in table 1. Figure 7 depicts the performance of all the models in terms of ROC curve.

\begin{figure}
\centering
\includegraphics[height=2in, width=2.5in]{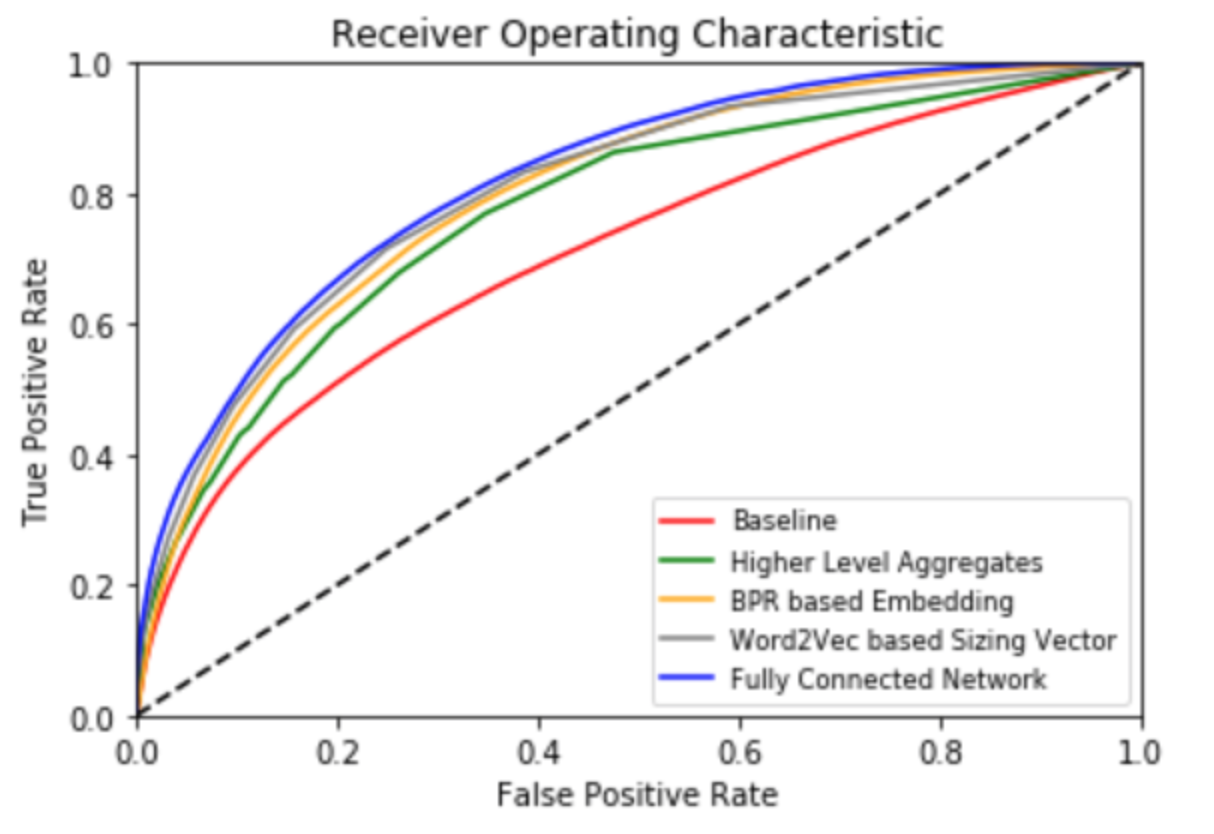}
\caption{Comparison of Model performance using ROC, which clearly show that Deep neural network based model as a winner}
\label{figure:architecture1}
\end{figure}

\begin{table}
\centering
\begin{tabular}{|l|l|l|l|}  \hline
\textbf{Model}                        & \textbf{AUC}  & \textbf{Precision} & \textbf{Recall} \\ \hline
Baseline                     & 71.6 & 65        & 18     \\ \hline
Higher Level Aggregates      & 80.1 & 69        & 32     \\ \hline
BPR based Embedding          & 81.9 & 69        & 34     \\ \hline
Word2Vec based Sizing Vector & 81.9 & 71        & 34     \\ \hline
Fully Connected Network      & 83.2 & 74        & 34     \\ \hline
\end{tabular}

\caption{Model Results}
\end{table}

\subsection{Results} \label{target-classifier}
In this section, the results of different models are discussed in a detailed manner (table 1). A baseline gradient boosted classifier created using basic user and product observable features resulted in 71.6\% AUC. To improve this baseline model, extensive feature engineering was performed which boosted AUC by 8\%.
Another 1.8\% increment in AUC was achieved by using the product embeddings learned using BPR. To further improve the precision, personalized sizing vectors at the user, brand level were used resulting in a 2\% increment in precision. These sizing vectors were learned using the skip-gram word2vec model. Further improvement in the AUC was difficult to achieve using the boosted classifier. Hence, a fully connected deep neural network model was trained using aggregated embeddings of product \& sizing along with engineered features as the input layer. This model out-performed all previous models with AUC, precision increased to 83.2\%, 74\% respectively. Figure 7 depicts the comparison between the ROC curves of all the models, with the deep neural network achieving the highest area under the curve.
    
    Some of the most important features turned out to be the product return score (product's lifetime return rate), cart size, payment type, user's lifetime returned quantity, sizing vectors, user's purchase frequency, product price, product embeddings. The first few features were discussed in Section 4.2. Apparently, these features turned out to be important from the model's perspective too.

\section{Experiments} \label{target-classifier}
In this section, the live experiments performed on one of the leading fashion platform are discussed. Experiments were conducted on three action items from the suggested ones in Section 1. First experiment comprised of applying personalized delivery charge which was different for all customers depending on their respective live cart. The dual model first predicts the return probability for a cart and then use this in a gradient boosted approach to identify the exact number of products that will be returned from that cart. Combining this result along with the business logic gives the final value of personalized delivery charge which varies from 0 to 149. An A/B test was performed on a live production environment for nearly 100K users. The conclusion derived from this experiment was that even though the number of orders reduced by 1.7\% in the test set as compared to the control set, the return percentage dropped even by a higher number (3\%). Further analysis indicated that 90\% of churned users had a high probability of return.
    
    The second experiment was to make product non-returnable by providing an additional coupon. The product-level model gave the return probability of each individual product in the cart. If this probability is high, we offer a small nudge of additional coupon and make that product non-returnable. This compensates for the reverse logistics or possible damage costs. The result from A/B test showed that 27\% of the users applied this option in the cart and marked 1.5 products (on an average) in their cart as non-returnable. This reduced the return rate in the test set by 4\% as compared to the control set.
    
    Third, Try \& Buy experiment was conducted where the users can try out their purchases at the time of delivery, keep what they like and return the rest on-the-spot. The hypothesis behind conducting this experiment was that it will reduce the reverse-logistics cost incurred during a return. Adoption of try \& buy feature was 40\% which reduced the return rate in the test set by 3.7\% as compared to the control set.

\section{Conclusion} \label{target-classifier}
E-tailers experience return-rate problem leading to increased costs and lower profit margins. To solve this, a novel approach is proposed in this paper to detect returns and implement certain actions items. Depending on the current cart configuration of a user, a hybrid dual-model is built using a deep neural network to detect returnable carts, products in real-time. Experiment results on action items show that accurate prediction of returns can lead to a reduction in the rate of return. As future work, we plan to apply this model on more action items which can further help in reducing the overall returns.



%
\bibliographystyle{IEEEtran}
\bibliography{sigproc}  

\begin{thebibliography}{10}
\providecommand{\url}[1]{#1}
\csname url@samestyle\endcsname
\providecommand{\newblock}{\relax}
\providecommand{\bibinfo}[2]{#2}
\providecommand{\BIBentrySTDinterwordspacing}{\spaceskip=0pt\relax}
\providecommand{\BIBentryALTinterwordstretchfactor}{4}
\providecommand{\BIBentryALTinterwordspacing}{\spaceskip=\fontdimen2\font plus
\BIBentryALTinterwordstretchfactor\fontdimen3\font minus
  \fontdimen4\font\relax}
\providecommand{\BIBforeignlanguage}[2]{{%
\expandafter\ifx\csname l@#1\endcsname\relax
\typeout{** WARNING: IEEEtran.bst: No hyphenation pattern has been}%
\typeout{** loaded for the language `#1'. Using the pattern for}%
\typeout{** the default language instead.}%
\else
\language=\csname l@#1\endcsname
\fi
#2}}
\providecommand{\BIBdecl}{\relax}
\BIBdecl

\bibitem{bonifield2010product}
C.~Bonifield, C.~Cole, and R.~L. Schultz, ``Product returns on the internet: A
  case of mixed signals?'' \emph{Journal of Business Research}, vol.~63, no.
  9-10, pp. 1058--1065, 2010.

\bibitem{article}
S.~Bandyopadhyay and A.~Paul, ``Equilibrium returns policies in the presence of
  supplier competition,'' \emph{Marketing Science}, vol.~29, pp. 846--857, 09
  2010.

\bibitem{doi:10.1108/03090561011032694}
\BIBentryALTinterwordspacing
L.~C. Harris, ``Fraudulent consumer returns: exploiting retailers' return
  policies,'' \emph{European Journal of Marketing}, vol.~44, no.~6, pp.
  730--747, 2010. [Online]. Available:
  \url{https://doi.org/10.1108/03090561011032694}
\BIBentrySTDinterwordspacing

\bibitem{Chen2009TheIO}
J.~Chen and P.~C. Bell, ``The impact of customer returns on pricing and order
  decisions,'' \emph{European Journal of Operational Research}, vol. 195, pp.
  280--295, 2009.

\bibitem{doi:10.1108/09600031011018055}
\BIBentryALTinterwordspacing
J.~B. Edwards, A.~C. McKinnon, and S.~L. Cullinane, ``Comparative analysis of
  the carbon footprints of conventional and online retailing: A “last mile”
  perspective,'' \emph{International Journal of Physical Distribution \&
  Logistics Management}, vol.~40, no. 1/2, pp. 103--123, 2010. [Online].
  Available: \url{https://doi.org/10.1108/09600031011018055}
\BIBentrySTDinterwordspacing

\bibitem{ma2016predictive}
J.~Ma and H.~M. Kim, ``Predictive model selection for forecasting product
  returns,'' \emph{Journal of Mechanical Design}, vol. 138, no.~5, p. 054501,
  2016.

\bibitem{toktay2001forecasting}
B.~Toktay, \emph{Forecasting product returns}.\hskip 1em plus 0.5em minus
  0.4em\relax INSEAD, 2001.

\bibitem{7385772}
A.~Canda, X.~Yuan, and F.~Wang, ``Modeling and forecasting product returns: An
  industry case study,'' in \emph{2015 IEEE International Conference on
  Industrial Engineering and Engineering Management (IEEM)}, Dec 2015, pp.
  871--875.

\bibitem{rendle2009bpr}
S.~Rendle, C.~Freudenthaler, Z.~Gantner, and L.~Schmidt-Thieme, ``Bpr: Bayesian
  personalized ranking from implicit feedback,'' in \emph{Proceedings of the
  twenty-fifth conference on uncertainty in artificial intelligence}.\hskip 1em
  plus 0.5em minus 0.4em\relax AUAI Press, 2009, pp. 452--461.

\bibitem{DBLP:journals/corr/abs-1301-3781}
\BIBentryALTinterwordspacing
T.~Mikolov, K.~Chen, G.~Corrado, and J.~Dean, ``Efficient estimation of word
  representations in vector space,'' \emph{CoRR}, vol. abs/1301.3781, 2013.
  [Online]. Available: \url{http://arxiv.org/abs/1301.3781}
\BIBentrySTDinterwordspacing

\bibitem{Urbanke2015PredictingPR}
P.~Urbanke, J.~Kranz, and L.~Kolbe, ``Predicting product returns in e-commerce:
  The contribution of mahalanobis feature extraction,'' in \emph{ICIS}, 2015.

\bibitem{mnih2008probabilistic}
A.~Mnih and R.~R. Salakhutdinov, ``Probabilistic matrix factorization,'' in
  \emph{Advances in neural information processing systems}, 2008, pp.
  1257--1264.

\bibitem{arora2016decoding}
S.~Arora and D.~Warrier, ``Decoding fashion contexts using word embeddings.''

\bibitem{hoyer2004non}
P.~O. Hoyer, ``Non-negative matrix factorization with sparseness constraints,''
  \emph{Journal of machine learning research}, vol.~5, no. Nov, pp. 1457--1469,
  2004.

\bibitem{Mikolov2013DistributedRO}
T.~Mikolov, I.~Sutskever, K.~Chen, G.~S. Corrado, and J.~Dean, ``Distributed
  representations of words and phrases and their compositionality,'' in
  \emph{NIPS}, 2013.

\end{thebibliography}
%
%
\end{document}